\def\year{2019}\relax
\documentclass[letterpaper]{article} %DO NOT CHANGE THIS
\usepackage{aaai19}  %Required
\usepackage{times}  %Required
\usepackage{helvet}  %Required
\usepackage{courier}  %Required
\usepackage{url}  %Required
\usepackage{graphicx}  %Required
\usepackage{clrscode}
\begin{document}
\title{TGE-viz : Transition Graph Embedding for Visualization of Plan Traces and Domains}
% \author{Paper ID: 252
% }

\author{Sriram Gopalakrishnan, Subbarao Kambhampati\\
Computing, Informatics, and Decision Systems Engineering\\
Arizona State University\\
699 S Mill Ave, Tempe, AZ 85281
}
\maketitle
\begin{abstract}
Existing work for plan trace visualization in automated planning uses pipeline-style visualizations, similar to plans in Gantt charts. Such visualization do not capture the domain structure or dependencies between the various fluents and actions. Additionally, plan traces in such visualizations cannot be easily compared with one another without parsing the details of individual actions, which imposes a higher cognitive load. We introduce TGE-viz, a technique to visualize plan traces within an embedding of the entire transition graph of a domain in low dimensional space. TGE-viz allows users to visualize and criticize plans more intuitively for mixed-initiative planning. It also allows users to visually appraise the structure of domains and the dependencies in it.
\end{abstract}

\section{Introduction}
One of the barriers to the adoption of automated planners is their usability. This is due to the amount of time and knowledge needed to interpret any output and interact with the planner. 
An area in which we can improve usability is plan trace and domain visualization. Current plan trace visualizations represent plans in a pipeline or linear sequence. 
% Such representations can give us information about timelines, or the ordering (maybe partial ordering) of actions.
If there is no complete ordering of actions, adjacent actions may have no immediate relationship or dependence. So the user would have to keep the effects of actions in mind, and connect it with a future action to realize the need for the prior action. Consequently, the user may have to parse the entire plan, before beginning to conceptualize about other possible plans. This is because the user needs to know about the dependencies across the plan. This high cognitive load often leads to mental fatigue in the user, which reduces the quality of plan criticism in mixed-initiative planning. Thus, it is important to present information in a visual and easy-to-parse (and recall) format. This would allow users to quickly generate alternate plans, or modify existing ones and compare them. 

Indeed, the Ecological Interface Design Principles \cite{eid_vicente1992ecological}, which helped set the standards for design in complex human-machine systems, require that the correct affordances (actions) are easily inferable to the operator.
To this end, we introduce TGE-viz, a visualization approach that uses ideas from graph embeddings to display the entire transition graph of a domain in $2$ dimensions. Graph embeddings are a popular method to reduce the vast amount of information in graphs by embedding their vertices to a continuous space to speed up analytics \cite{graphSurvey}. Lower dimensional embeddings allow humans to be involved in the analysis. We can intuitively see structures (clusters and shapes) and relative distances, which can be used to augment any automated analysis. It is this intuitive understanding and human insight that we hope to bring into mixed-initiative planning.

We will first discuss some relevant parts of graph theory which will be used in this paper. Then we will discuss the TGE-viz algorithm and analyze some experimental results about the embeddings in the Logistics and Barman planning domains. This is followed by the discussion of the user interface for plan trace visualization, and interacting with the automated planner. Our visualization is then compare to existing techniques in the related work section.

\section{Background}\label{sec:intro}
We assume the reader is familiar with STRIPS style planning, for more details on it, we refer the reader to \cite{stripsIntro}. This section will focus on the graph theory concepts necessary for TGE-viz. 

In graph theory, Hopcount is one of the metrics to measure distance between nodes. Hopcount between two nodes is number of hops or links on the shortest path between the two nodes. 
\begin{equation}\label{eq:1}
H_{A \rightarrow B} =  \min_K (P_{A \rightarrow B}(k))
\end{equation}
where $H_{A \rightarrow B}$ is the hopcount between A and B, and $P_{A \rightarrow B}(k)$ is the $k^{th}$ path from A to B. 

% < if pages permit IMAGES OF HOPCOUNT , node with closeness, and a graph with high average closeness (make with google slides/drawings)>

Closeness of a node $n_i$ is the average hopcount from this node to all other nodes. Typically the reciprocal of the total hopcount from $n_i$ is used \cite{graphMetrics}

\begin{equation}\label{eq:2}
C_{n_i} =  \frac{1}{\Sigma_{n_j \in N/\{n_i\} } H_{n_i \rightarrow n_j}}
\end{equation}

Closeness indicates how tightly coupled a node is with other nodes. So the average closeness $C_g$ over all nodes in the graph is used as a measure of the closeness of the overall graph. The Radius is another helpful metric. First we define the eccentricity of a node as the longest hopcount to any other node. The radius of a graph is the minimum node eccentricity over all nodes. We will discuss the relationship of these graph metrics and the quality of TGE-viz embeddings. 

\section{Problem Formulation }\label{sec:problem_form}

The objective of TGE-viz is that given a domain specification, find embeddings for all grounded fluents and grounded actions such that the k-nearest neighbors of the graph over grounded actions and fluents are preserved. This optimization metric is used to capture the structure and relationships of the domain. Henceforth, we shall refer to the graph connecting the actions and fluents of the domain as just the transition graph. 

The input to the problem is the set of all grounded actions $G_a$. Each $g_a$ is the triple $(P,a,E)$, where $P$ is the set of precondition fluents (each of which is a string), $a$ is a string that represents the grounded action, and $E$ is the set of effects. All the nodes in the domain are represented by $\tau = V \bigcup A$, where A is the set of all action strings from $G_a$, and $V$ is the set of all fluents in the domain.

%the following lines can be cut
The transition graph is built by adding edges between every action $a \in A$ and fluent $v \in V$ that is a precondition or effect (includes the delete list) in $G_a$. The output is the set of tuples $(t_i,e_i)$ where $t_i$ is the $i^{th}$ term of $\tau$, and $e_i$ is its embedding. The embeddings are optimized such that neighboring nodes in the graph are closer together, and non-neighbors are further away.
How the optimization metrics are concretely defined depends on the graph embedding algorithm used which we will discuss shortly.

\section{Embedding a Planning Domain }\label{sec:embedding_main}

Graph embedding is a rich field with many algorithms that have their relative merits. For this work we experimented with  Multi-Dimensional Scaling (metric and non-metric MDS), Locally Linear Embedding, Isomaps, and Spectral clustering. The aforementioned algorithms were run with the implementations in Sci-kit python library \cite{scikit-learn}.
We found that what worked best was a $\it{variation}$ of Force-based graph embedding \cite{ge_forceBased}, which is one of the original techniques of graph embedding, and was intended to provide helpful visualizations of graphs (often called a graph $drawing$ algorithm). There are many variants within this class of graph drawing
algorithms as compared in \cite{comparisonGraphAlg}. We chose to code a $\emph{variant}$ of the Fruchterman-Reingold algorithm for its simplicity, speed and scalability for large transition graphs. 

\subsection{TGE-viz Graph Embedding Algorithm}\label{sec:algorithm}
The high-level algorithm for TGE-viz is in Figure \ref{TGE_viz_algo}, which builds the transition graph and then updates the embeddings.
%-----------------------------------------
\begin{figure}[!ht]
\begin{codebox}
\Procname{$\proc{graphEmbedder}(A,\tau,iterations = 1500)$}
\li G $\leftarrow$ BuildGraph(A)
\li $\backslash \backslash$ Initialize the embeddings between 0,100 in 2d
\li E $\leftarrow$ InitEmbeddings($\tau$,0,100,2)
\li \For $i \in range(0,iterations)$  do
\li \ \ \ \ E $\leftarrow$ UpdateEmbeddings(E,G,$\tau,\alpha$=1.0)
\li $return$ E
\End
\end{codebox}
\caption{High-level algorithm for embedding a transition graph}
\label{TGE_viz_algo}
\end{figure}
%----------------------------------------------
\begin{figure}[!ht]
\begin{codebox}
\Procname{$\proc{BuildGraph}(G_a)$}
\li G $\leftarrow$ $\emptyset$
\li \For each $g_a \in G_a$  do
\li \ \ \ \ $\backslash \backslash$ each action is comprised of a set of preconditions,
\li \ \ \ \ $\backslash \backslash$ the actionID, and set of effects
\li \ \ \ \ (P,$a$,E) $\leftarrow g_a$ 
\li \ \ \ \ \For each $v \in$ $P \bigcup E$  do
\li \ \ \ \ \ \ \ \ G = G $\bigcup$ \{(a,v)\}
\End
\end{codebox}
\caption{Building the graph for embedding}
\label{build_graph}
\end{figure}
%------------------------------------------
The crux of the algorithm is in how the embeddings are updated from their initial random positions. This is in Figure \ref{update_embeddings}. 
\begin{figure}[!ht]
\begin{codebox}
\Procname{$\proc{UpdateEmbeddings}(E,G,\tau,\alpha=1.0)$}
\li $\backslash \backslash$ E is the current set of embeddings
\li $\backslash \backslash$ create a copy into B (base set)
\li B = copy(E)
\li E $\leftarrow$ $\emptyset$
\li $\backslash \backslash$ randomly select log($|\tau|$) number of terms for repulsion
\li $\backslash \backslash$ more cost-efficient that repelling from all non-neighbors
\li R $\leftarrow$ RandomSelect($\tau, log(|\tau|$))
\li \For each $w \in \tau $   do
\li \ \ \ \ attrForce $\leftarrow$ $\vec{0}$ 
\li \ \ \ \ repelForce $\leftarrow$ $\vec{0}$
\li \ \ \ \ $\vec{w}$ $\leftarrow$ GetCurrentEmbedding(w,B)
\li \ \ \ \ \For each $n \in $ neighbors(G,w) do
\li \ \ \ \ \ \ \ \ $\vec{n} \leftarrow$ GetCurrentEmbedding(n,B)
\li \ \ \ \ \ \ \ \ attrForce = attrForce + $(\vec{n} - \vec{w})$
\li \ \ \ \ \For each $r \in $ R do
\li \ \ \ \ \ \ \ \ $\vec{r} \leftarrow$ GetCurrentEmbedding(r,B)
\li \ \ \ \ \ \ \ \ $\backslash \backslash$ repulsion is inv proportional to distance
\li \ \ \ \ \ \ \ \ repulsion = $\frac{|\tau|}{log(|\tau|)} * \frac{1}{\vec{r} - \vec{w}}$
\li \ \ \ \ \ \ \ \ repelForce = repelForce + repulsion
\li \ \ \ \ $\vec{m} \leftarrow $ attrForce - repelForce
\li \ \ \ \ $\vec{d}$ = min($(\vec{m} , \vec{m}/|\vec{m}|)*\alpha)\backslash \backslash$ limit by learning rate
\li \ \ \ \ $ E \leftarrow E \bigcup \{(w, \vec{w} + \vec{d})\} \backslash \backslash$ store updated embedding
\li $return$ E
\End
\end{codebox}
\caption{Process to update the embeddings}
\label{update_embeddings}
\end{figure}
%----
Each node is pulled towards it's neighbors with a force proportional to the distance between them. Each node is also repelled from non-neighbors $R$ by a force inversely proportional to the distance between them. For each iteration, we randomly select \emph{only} $log(|\tau|)$ nodes for repulsion (into the set $R$) to speed up computation. At each iteration, the node can only move up to a set max-distance $\alpha$. 
%---We can delete the following lines
More even distribution of embeddings can be achieved by repelling from all non-neighbors in every iteration, rather than a random set of size $log(|\tau|)$. However, this will not scale well with very large number of nodes, and it is not necessary to produce helpful visualizations of the domain.

If the reader is interested in visualizing the update process of the TGE-viz algorithm, we have uploaded an anonymized video\footnote{Video of embeddings update https://youtu.be/WB8XvJI01SA}(see footnote). 

This embedding algorithm works well for a lot of graph configurations, and the degree of usefulness of the visualization depends on the closeness of the graph as well as its radius. We tested our approach to analyze that, as follows.

\section{Testing Methodology}
We ran TGE-viz on various configurations of two representative domains, Logistics and Barman. The standard logistics domain allows airplanes to fly to all airport locations. This allows for very little variability in the transition graph structure. So, we added constraints to let certain airplanes only access specific airport locations (specific cities), and not all airport locations.
We chose the Logistics domain as it is representative of the class of transition graphs where there is more structure and separation between different parts of the domain as evidenced by the low average-closeness score and larger radius in Table \ref{tbl:graphMetrics}. 

On the other hand, the Barman domain has transition graphs that have high average closeness score. This is because there are significantly more actions that connect a fluent to another, and each fluent is connected to a much larger ratio of the total set of fluents by very short hopcounts. As a result, the closeness is higher and radius of the graph is much smaller as seen in Table \ref{tbl:graphMetrics}. 

Barman serves to highlight the kind of domains where the embeddings using TGE-viz, will be less separated. As we will see in the results, even in Barman we can see the structure in the domain.

For details on how the various domain were configured, please see the supplementary material for the PDDL domain and problem files. In short, for Logistics domain we increased the connectivity between cities by adding airplanes that connect them. In Barman, we increased the number of cocktails and shot-glasses (cups). 
\section{Experimental Results and Analysis for Graph Embeddings}
\begin{figure}[!ht]
% \centering
\includegraphics[width=3.3in]{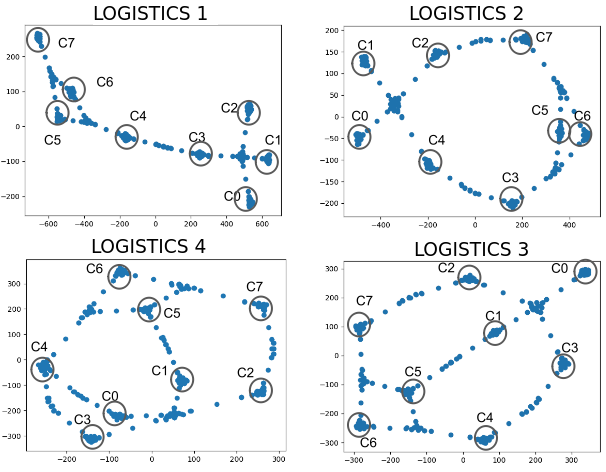}
\caption{Logistics Domain Embeddings}
\label{fig:logisticsEmbeddings}
\end{figure}

\begin{figure}[!ht]
% \centering
\includegraphics[width=3.3in]{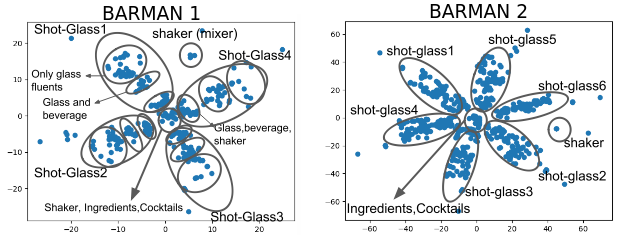}
\caption{Barman Domain Embeddings}
\label{fig:barmanEmbeddings}
\end{figure}

The embeddings for various configurations of Logistics and Barman are presented in Figures \ref{fig:logisticsEmbeddings} and \ref{fig:barmanEmbeddings}, followed by their graph metrics in Table \ref{tbl:graphMetrics}
%--------------------------------
%--------------------------------------
%--------------------------------------

\begin{table}[!ht]
  \begin{center}
    \caption{Graph Metrics}
    \label{tbl:graphMetrics}
    \begin{tabular}{|c|c|c|} % <-- Alignments: 1st column left, 2nd middle and 3rd right, with vertical lines in between
    \hline
      \textbf{Name} & \textbf{Average Closeness} & \textbf{Radius}\\
%       $\alpha$ & $\beta$ & $\gamma$ \\
      \hline
      Logistics 1 & 0.067 & 18\\
      \hline
      Logistics 2 & 0.079 & 20\\
      \hline
      Logistics 3 & 0.086 & 16\\
      \hline
      Logistics 4 & 0.090 & 16\\
      \hline
      Barman 1 & 0.429 & 2\\
      \hline
      Barman 2 & 0.41 & 2\\
      \hline
    \end{tabular}
  \end{center}
\end{table}

For the Logistics embeddings in Figure\ref{fig:logisticsEmbeddings}, $C1, C2,...C7$ represent the fluents and actions related to transportation within each of those cities. The embeddings between cities represent the transportation of a package between them. As we add more airplanes and possible actions to transport packages between cities, the graph embeddings update to reflect this in their structure. The clean separation and structure is because the Logistics graphs have a low closeness score (think degree of connectivity), and the radii are larger. This will result in the TGE-viz algorithm spreading the fluents and actions wider in the embedding space. This makes it easier to extract information from the visualization.

On the other hand, Barman domain has a very high closeness score and small radius. So the embeddings are not as spread out as in Logistics and thus makes it less clean (separated) for visualization and ascribing meaning to clusters and paths in the embedding space. Even in the densely connected Barman domain we can clearly see separation in the embeddings as highlighted in figure \ref{fig:barmanEmbeddings}. The fluents and actions related to each shot-glass (cup) form protruding prongs from the center. The fluents and actions related to mixing ingredients in the shaker and making cocktails are in the center. Very weakly connected nodes that have few edges like $clean\_shot1$ and $clean\_shaker$ are pushed to the periphery. 

Hence, one way to determine if a transition graph can be easily visualized with TGE-viz is to use the closeness and radius scores. Lower closeness and higher radius scores are better. For very large graphs, it maybe computationally cheaper to just see the result. Using such graph embeddings, we now present the first version of our user interface that lets a user visualize plan traces with respect to the entire domain structure, and interact with the planner through it.

\subsection{Mixed-Initiative User Interface with TGE-viz}
% Show UI developed in pygame. Explain the options and the zoom capability. Apologize for the coloring, and mention it will be corrected in future versions to have different shapes. 

\begin{figure}[!ht]
% \centering
\includegraphics[width=3.3in]{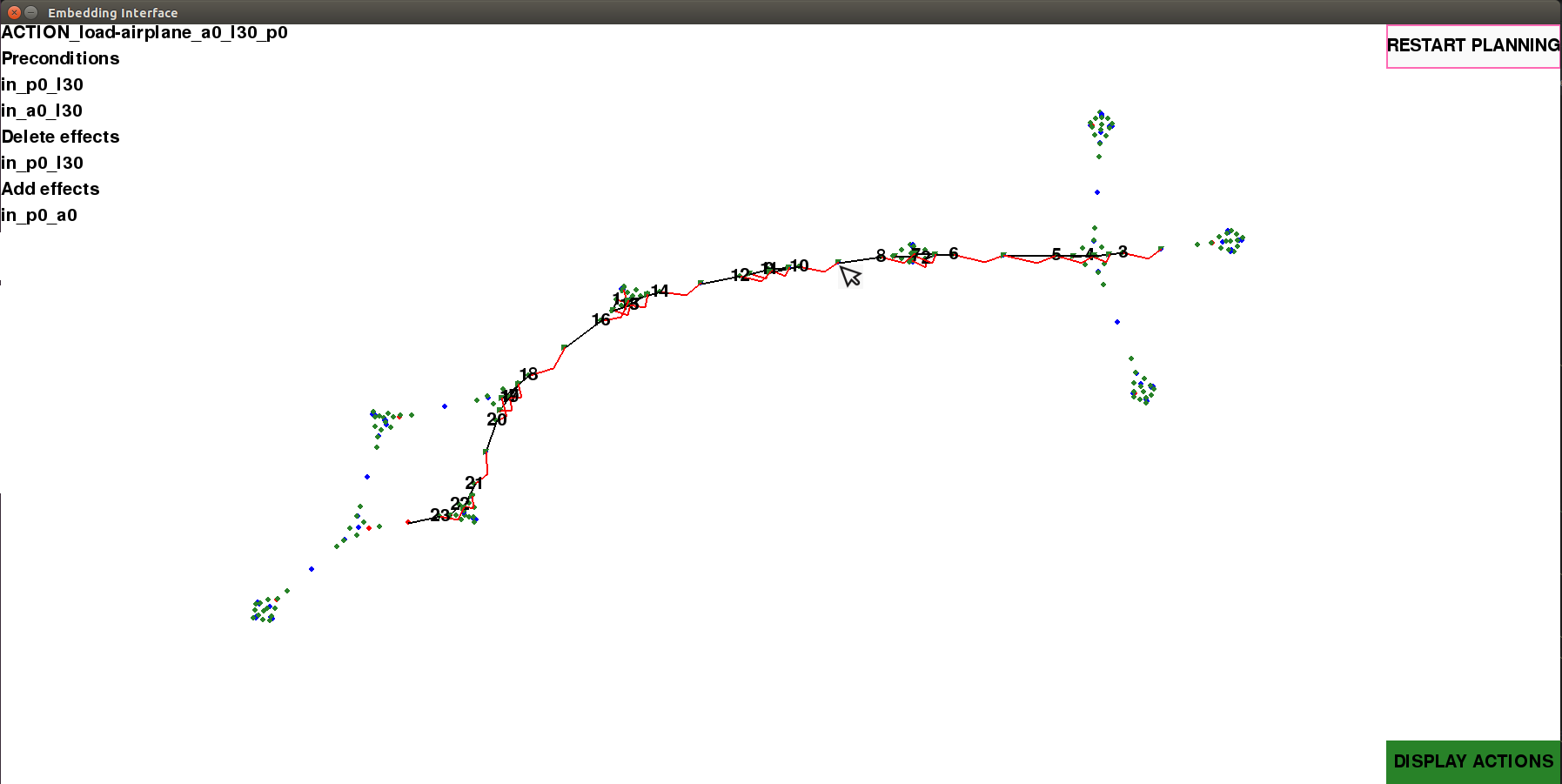}
\caption{ Plan trace in modified logistics with TGE-viz for the goal of delivering the package to city 6 location 3}
\label{fig:pygame}
\end{figure}

We developed the proof-of-concept user interface in PyGame \cite{pygame}. Figure \ref{fig:pygame} is a screenshot capture of our interface. The interface first displays all the fluents and actions in the domain. The actions are green, the current state fluents are red, and other fluents are displayed in blue. Currently, the initial state is fed in through a $problem.pddl$ file in the standard pddl format. 
The user can choose to turn off the display of actions by clicking the button in the bottom right as there typically tend to be a lot more actions than fluents. When the user hovers the mouse over a node, its information is displayed in top left. When the user clicks a fluent, the interface calls FastDownward planner \cite{helmert2006fast} in the background, setting the clicked fluent as the goal. The resulting plan is displayed on the embeddings. A red line connects each action to the preconditions consumed, and a black line to the effects produced. The user can continue planning from the resultant state or restart using the restart button on the top right. 

\begin{figure}[!ht]
% \centering
\includegraphics[width=3.3in]{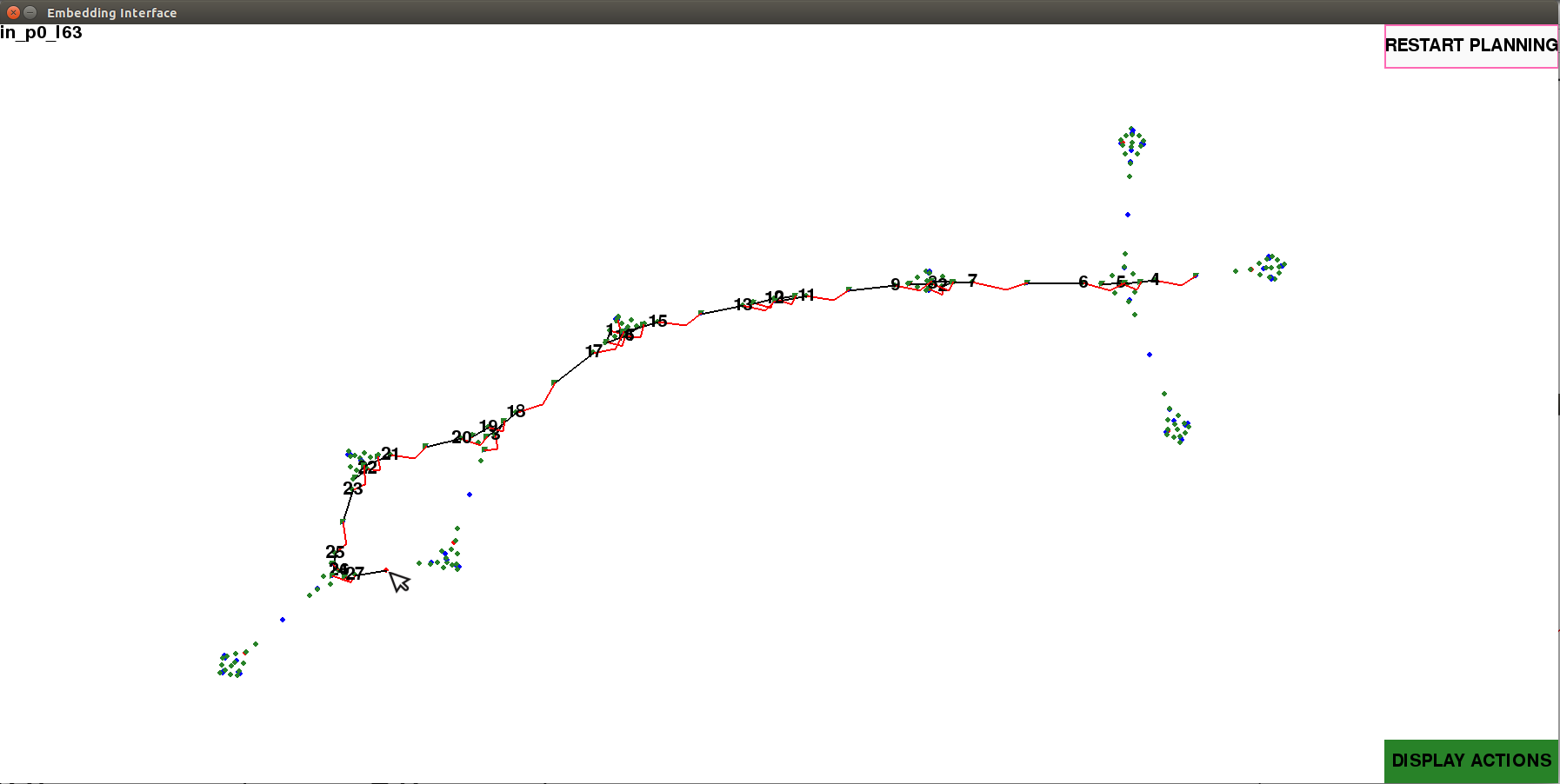}
\caption{ Alternate plan trace in modified logistics with TGE-viz for the goal of delivering the package to city 6 location 3}
\label{fig:pygame_alt}
\end{figure}

The plan traces are thus visualized in the context of the entire graph. If the user wishes to try other plans, they can click appropriate subgoals to guide the planner through a different path. In the example in Figure \ref{fig:pygame}, the user can spot and try an alternative plan traces as shown in Figure \ref{fig:pygame_alt}. Other smaller features of our interface, such as zoom (magnification), are discussed in section 1 of the supplementary material.
%--------------------------------
%--------------------------------------
%--------------------------------------
\section{Related Work }
\begin{figure}[!ht]
% \centering
\includegraphics[width=3.3in]{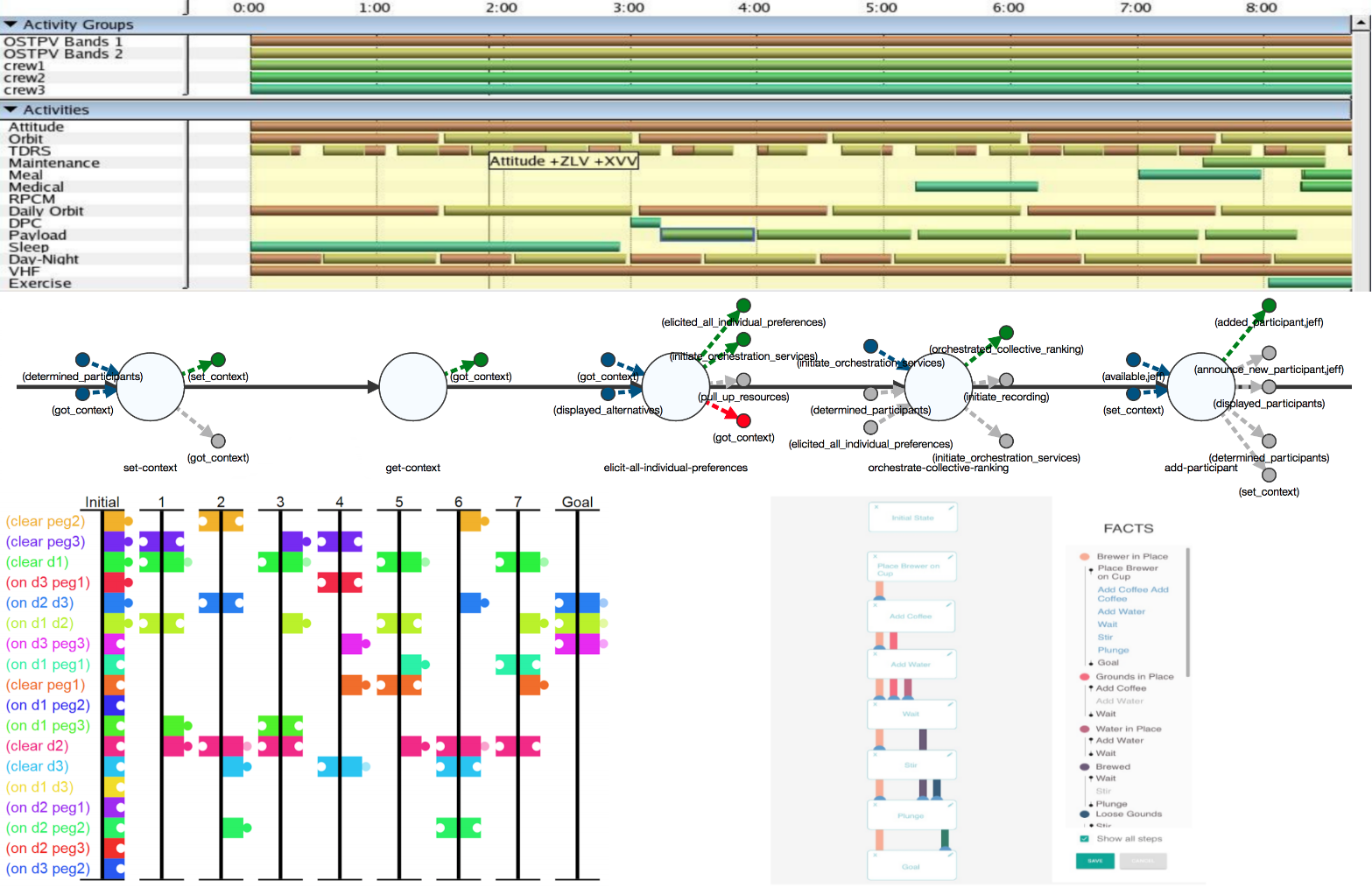}
\caption{ Existing Visualization For Planners. Clockwise: SPIFe, Fresco, Conductor, WEBPLANNER}
\label{fig:existingViz}
\end{figure}

Existing plan trace and domain visualizations present information in a sequential manner with no notion of the rest of the domain. Examples of such representations are Conductor \cite{bryce2017situ}, MAPGEN \cite{mapgen}, SPIFe \cite{spife_clement2010spatial}, Fresco \cite{chakraborti2017visualizations} and Webplanner\cite{webplanner_magnaguagno2017}. These can be seen in Figure \ref{fig:existingViz}.

%--------------------------------
%-----------------------------
Fresco \cite{chakraborti2017visualizations} tries to remedy this by displaying top-k plans. However, it is still left to the user to painstakingly parse the actions of each plan trace of the top-k plans to determine the differences. In comparison, our plan trace visualization of alternative traces in Figure \ref{fig:pygame} and Figure \ref{fig:pygame_alt} allow the user to immediately see where the differences are, and parse the information with less cognitive load.

\section{Conclusion and Future Work }
TGE-viz allows visualization of plan traces in the context of the transition graph of the entire domain. This enables the human in mixed-initiative planning to intuitively and quickly analyze plan traces overlayed on it. Alternate plans can be formulated easier, and our interface provides an easy way to interact with the automated planner. We think that such interfaces that reduce the cognitive load of the user will help the adoption of automated planners. We will experiment with other algorithms for drawing graphs, some of which are discussed in section 3 of the supplementary material.
In the next version of our interface,  we would like to display time and other costs associated with the actions in the plan traces displayed on the embeddings.

\newpage
\bibliographystyle{aaai.bst}
\bibliography{icaps2019_PlanViz.bib}

\newpage
\section{SUPPLEMENTAL MATERIAL}
% \def\year{2019}\relax
% %File: formatting-instruction.tex
% \documentclass[letterpaper]{article} %DO NOT CHANGE THIS
% \usepackage{aaai19}  %Required
% \usepackage{times}  %Required
% \usepackage{helvet}  %Required
% \usepackage{courier}  %Required
% \usepackage{url}  %Required
% \usepackage{graphicx}  %Required
% \usepackage{clrscode}
% \frenchspacing  %Required
% \setlength{\pdfpagewidth}{8.5in}  %Required
% \setlength{\pdfpageheight}{11in}  %Required
% %PDF Info Is Required:
%   \pdfinfo{
% /Title (2019 Formatting Instructions for Authors Using LaTeX)
% /Author (AAAI Press Staff)}
% \setcounter{secnumdepth}{0}  
%  \begin{document}
% % The file aaai.sty is the style file for AAAI Press 
% % proceedings, working notes, and technical reports.
% %
% \title{SUPPLEMENTAL for TGE-viz : Transition Graph Embedding for Visualization of Plan Traces and Domains}

% % \author{Paper ID: 252
% % }

% \maketitle

% \section{TGE-viz for domain engineering}
% Domain Visualization:
% ANECDOTE Error in the structure connecting city 1 and city 5, the structure
% was disconnected. Show it. The error was that while copy pasting the location for airports I replaced the existing one, and did not add the line to have the airplane fly from city 1 to 5. As a result, the cluster of
% action and fluents was orphaned from the others. This clear structural defect pops up in the embedding. 

\section{1. Additional User Interface Functionality }

\subsection{1.1 Zoom In/Out }
\begin{figure}[!ht]
% \centering
\includegraphics[width=3.3in]{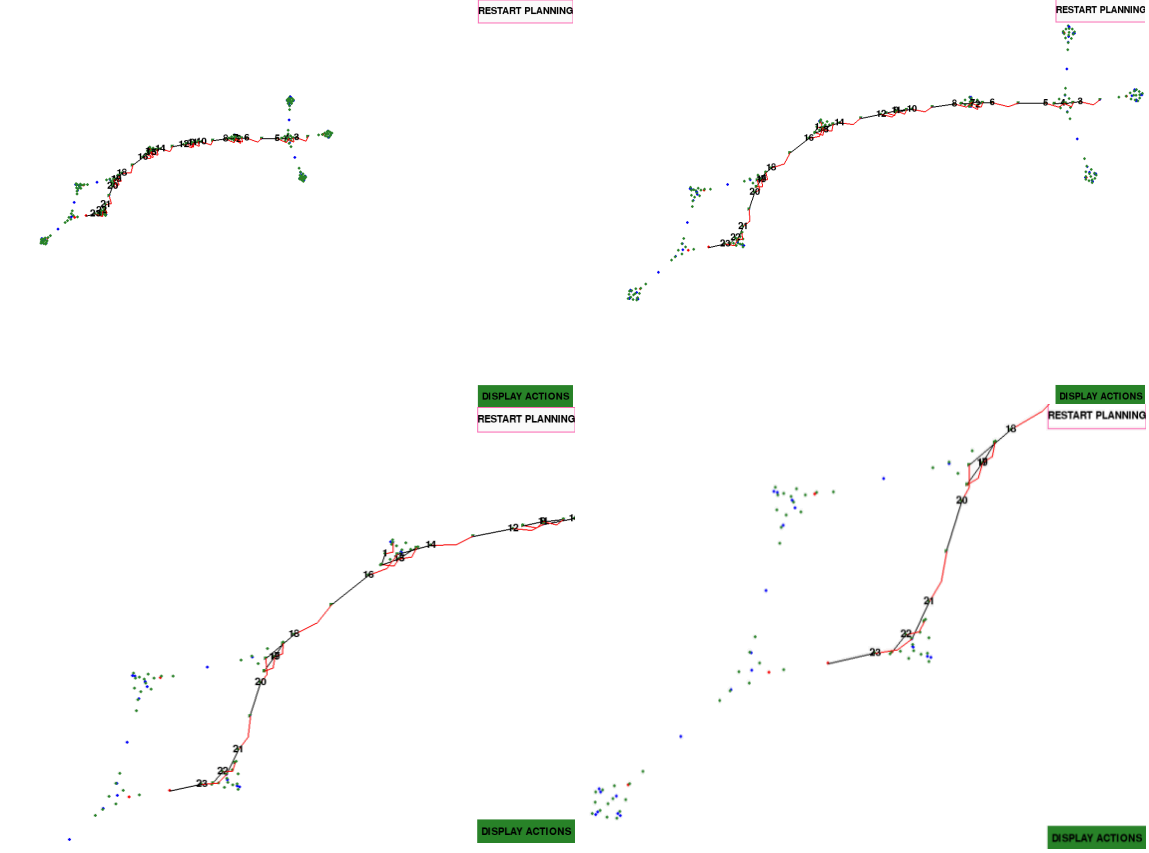}
\caption{Magnification in TGE-viz interface}
\label{fig:zoom_pygame}
\end{figure}

Since the embedding space can be large, and there can be subsets of nodes close together, we decided to prioritize support for the zoom (magnification) functionality. The interface allows zoom by the scroll wheel on a mouse, and magnifies based on the position of the cursor as shown in Figure \ref{fig:zoom_pygame}
%--------------------------------------
%--------------------------------------

\subsection{1.2 Visualization of Path of Multi-valued Fluents}
\begin{figure}[!ht]
% \centering
\includegraphics[width=3.3in]{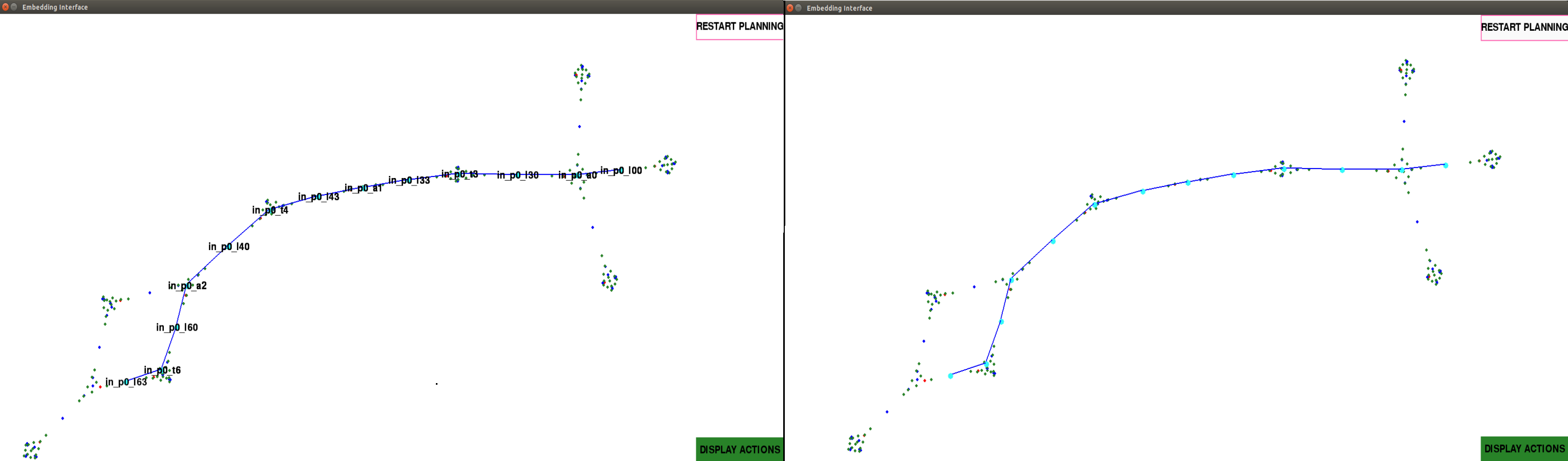}
\caption{Visualization of a plan trace with only the trajectory of the \emph{multi-valued} goal fluent (with and without labeling)}
\label{fig:fluent_path}
\end{figure}

If the domain was described with mutli-valued fluents (introduced in PDDL 3.1 \cite{pddl31_def}), then we can display the plan trace by filtering/simplifying it to only display the trajectory of changes to the goal fluent if it is a multi-valued fluent. This is already supported in our interface, see Figure \ref{fig:fluent_path} for an example in the logistics domain in which we treat the location of the package as a multi-valued fluent. It goes through values such as $in\_p0\_l33$ (package0 in location 33) and $in\_p0\_a1$ (package0 in airplane1). Note that the logistics domain does not define the location of a package as a multi-valued fluent. We are just presenting the package location as an example of one, since it does behave similarly. Such a plan trace visualization of only the goal fluent(s) further reduces the amount of information displayed to a useful subset. This in turn can make analysis less taxing. 

%--------------------------------------
%--------------------------------------
\subsection{1.3 Toggle Display of Static Fluents}
By default we choose to not embed or display static fluents (fluents that never change value) such as $airplane\_a1$ or $inCity\_l10\_c10$. We chose to do so because we wanted to reduce the informational overload to the user by cluttering the visualization. Since this is information that never changes state, we thought it was good to prune away from the embeddings visualization. We do realize that static fluents maybe used by the actions as preconditions. This information can just be displayed on demand when the user hovers over the action embedding, rather than clutter the display and increase the cognitive load.
We do support the embedding and display of static fluents if needed. It may help give semantic meaning to regions of the embedding space, and we hope to test this with human studies in future work.

\section{2. Debugging Domain Errors}
\begin{figure}[!ht]
% \centering
\includegraphics[width=3.3in]{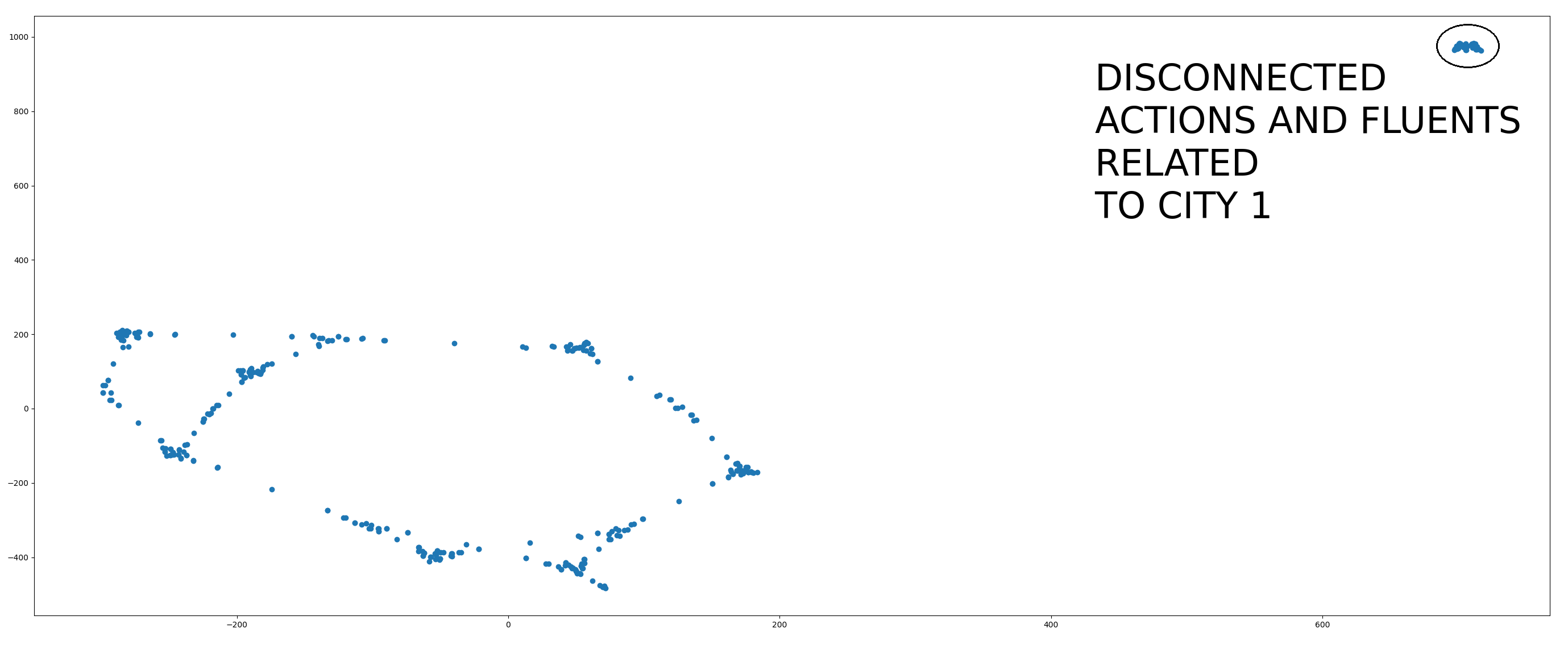}
\caption{Erroneous domain definition for Logistics 4 clearly caught in TGE-viz embeddings}
\label{fig:debug_embeddings}
\end{figure}

One unexpected use of our visualization that we discovered, was using TGE-viz to catch faulty domain descriptions.
During the testing phase of this work, we saw a problem through an unexpected set of embeddings. The embeddings of the fluents and actions related to transportation to $city\_1$ were completely separated from the rest of the embeddings as shown in Figure \ref{fig:debug_embeddings}. Using the visualization of the transition graph, we saw what the actions and fluents in the separated cluster were, and compared them to what we expected them to be closer to. We immediately noticed that the $fly \space airplane$ actions between the airport location of $city\_1$ and any other airport was missing. This helped us find the specification error that caused it, which was an accidentally deleted fluent that defined an airport location in $city\_1$. This occurred as we were trying to modify and scale up the domain by cutting and pasting fluents. Thus a possible additional use of TGE-viz embeddings is for debugging during domain engineering. Ofcourse, not all errors are as obvious as the case we pointed out. What we argue is that unexpected relative positions of fluents and actions, or any unexpected asymmetries in the TGE-viz embeddings can be used to quickly identify errors. This technique of debugging with visualizing the embeddings could be especially helpful when engineering a large domain. It can supplement the exhaustive testing method of running a large test suite of problems. The latter would require parsing the individual failure cases to determine the common thread of the errors, which is a heavier cognitive load on the user.

%--------------------------------------
%--------------------------------------

\section{3. Alternative Graph Drawing Algorithms and Analysis}

In this work we used a variant of the 
Fruchterman-Reingold (FR) algorithm \cite{ge_forceBased} which is a $Force\space Directed$ graph drawing algorithm. The computational complexity for the basic version of the FR algorithm is $O(|V|^2 + |E|)$ \cite{comparisonGraphAlg}. Our version is $O(|V|log(|V|) + |V||E|)$, since we only choose to repel from $log(|V|)$ terms every iteration. The second term is $|V||E|$ because in each iteration we update all the nodes, as opposed to one node at a time in the original FR algorithm. The real computational benefit of our approach is the parallelizability. This comes from the fact that we keep a fixed (base) set of embeddings each iteration, and update the embeddings for all nodes with respect to this. In the basic FR algorithm, each embedding update of a node affects the update of the next node. This tying of the updates of successive nodes within an iteration prevents the base FR algorithm from being parallelized. The parallelization options for TGE-viz is discussed more in the next section.

A big difference between the base FR algorithm and our version was to not limit the area of the embedding, which the FR algorithm uses as a parameter to scale the attractive and repulsive forces. We didn't find it necessary to convolute the algorithm with an additional parameter that would need tuning. One can always rescale the final embeddings (as we do) before displaying them in the interface. One other difference is that in the FR algorithm, the attraction force is proportional to the $distance^2$ between the neighboring nodes. We found it better in our experiments to just have it proportional to the distance, as the former ($distance^2$) resulted in embeddings that were not well spread, and distances within a cluster were too close, so the local differences and relationships were less apparent.

An optimization trick that we found worked quite well is as follows;  Replace the attractive forces with jumping the position of a node to the halfway point between the current embedding, and the centroid point defined by the all of the neighbor's embeddings. We call this $Half-Jumping$. Adding this change helps the convergence speed (although by not much). More importantly, we observed that this change helps the algorithm get past suboptimal configurations better than using force directed attraction. To deal with suboptimal configurations in the original FR algorithm, one might have to tune the $temperature$ parameter (which affects the size of the updates), and go through a cooling phase (as is done in simulated annealing). We find our trick of $Half-Jumping$ avoids the need for tuning the $temperature$ and a cooling phase. Note that the repulsion forces are still kept.  

A different approach to drawing a graph is to minimize the total energy of the system. The energy of the system is defined by the chosen graph theoretic distance(configurable) between nodes. The Kamada-Kawaii (KK) algorithm \cite{kamada1989algorithm} is the representative example of this other approach. The KK algorithm takes longer with a time complexity of $O(|V|^2log|V| + |V||E|)$. We have not yet explored the energy minimization approach to graph drawing, outside of the cursory time complexity analysis. Such energy minimization approaches have been criticized for their poor scalability \cite{springEmbeddersComparison} since they require the computation of the shortest path between all pairs. We think it may still be a worthwhile effort to compare the quality of embeddings produced by energy minimization algorithms. There maybe certain domain graphs for which the energy minimization approach works better.

\section{4. Scalable parallelization of TGE-viz}

A benefit of the force-directed embeddings approach is the ability to extensively parallelize the algorithm. The main step of TGE-viz is updating the embeddings. This step can be split up amongst parallel computational processes by dividing the terms in $\tau$ into as many parts as there are processes available. Each process only needs the repulsion set for that iteration ($log(|\tau|)$ number of embeddings), and the information for the nodes (embedding and neighborhood) that the computational process is handling. Once an iteration of the update step is completed, we merge the embedding updates to form the fixed (base) set of embeddings for the next iteration. This way of computing fits nicely into the map-reduce framework of computation \cite{mapreduce_dean2008}, and allows us to scale to very large domains. This will be a topic of research for extensions on this work.

\section{5. Visualizing search heuristics}

Other visualizations explored for automated planners include visualization of the plan search space to compare heuristics. This was done in $Webplanner$ \cite{webplanner_magnaguagno2017}.
%--------------------------------------
\begin{figure}[!ht]
% \centering
\includegraphics[width=3.3in]{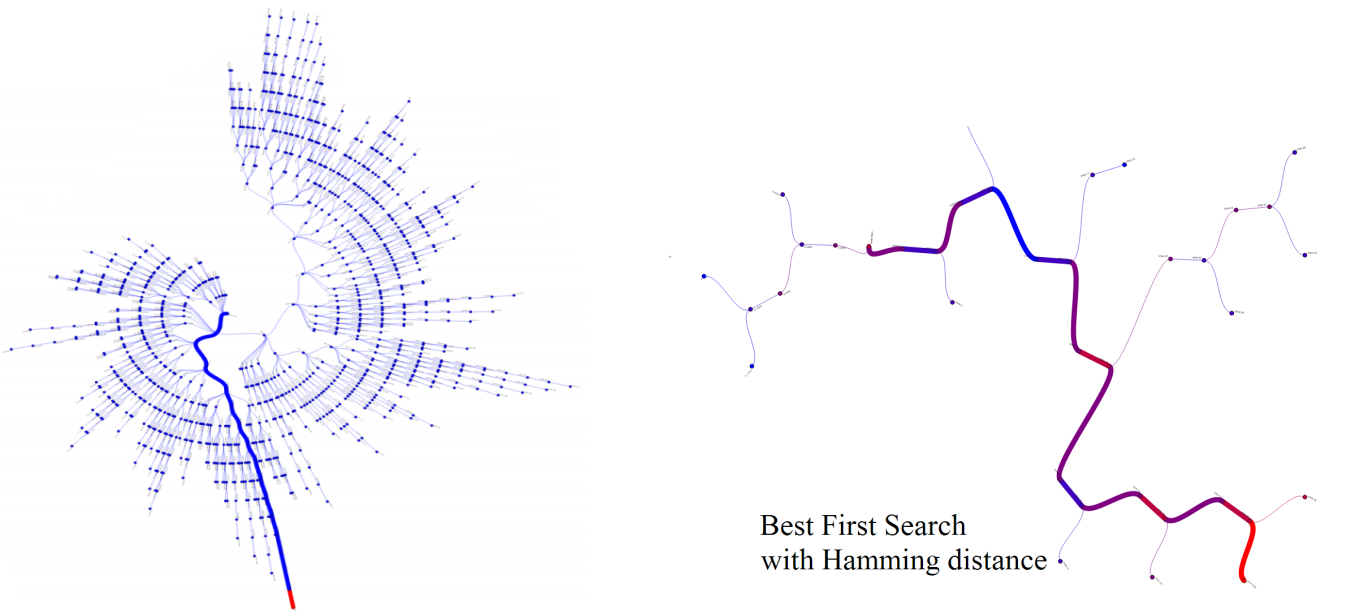}
\caption{Heuristic search process visualization in Webplanner}
\label{fig:search_heuristics_viz}
\end{figure} 
%--------------------------------------
They represented the search space as cartesian trees and radial graphs as shown in the bottom left of \ref{fig:search_heuristics_viz}. These can be used to see the search process of heuristics. However, for every new search, the user must look at each of the large number of nodes to see what they are. There is no persistence or meaning to the positions. On the other hand if the most successful or closest (by heuristic value) search paths were overlayed on the TGE-viz embedding, the user can see how the the domain was being explored by a heuristic. We have yet to implement this feature and may do so in a future version.

\section{6. Embeddings as Heuristics}

If the domain is described with multi-valued fluents (similar to state variables), then the distance between the current and goal fluent embeddings can be used as a heuristic in the plan search. We tested a proof of concept planner with this heuristic and the results are promising but currently far from being competitive with Fastdownward \cite{helmert2006fast} in the standard planning domains. To use graph embeddings as a heuristic, we need to compute embeddings in higher dimensional spaces. The distances are more informative in higher dimensions, but we will not be able to visualize them. Once the embeddings are computed, the $Embedding\space Distance\space  Heuristic$ can be used for all problems in the defined domain. So the cost of computing the embeddings is amortized over the set of all problems.

\section{7. Domain Configurations for Experiments}

The domain and problem $.pddl$ files used for the experiments in this work are contained in the supplemental material. If you cannot access it, and would like them, please contact us.

% \subsection{All Logistics Domain file}
% \subsection{Logistics 1 Problem file}
% \subsection{Logistics 2 Problem file}
% \subsection{Logistics 3 Problem file}
% \subsection{Logistics 4 Problem file}
% \subsection{Barman Domain file}
% \subsection{Barman 1 Problem file}
% \subsection{Barman 2 Problem file}

\bibliographystyle{aaai.bst}
\bibliography{icaps2019_PlanViz.bib}

\end{document}